# Budgeted Learning of Naïve-Bayes Classifiers

Daniel J. Lizotte     Omid Madani     Russell Greiner

Department of Computing Science
University of Alberta
Edmonton, AB T6G 2E8 Canada
{dlizotte, madani, greiner}@cs.ualberta.ca

## Abstract

There is almost always a cost associated with acquiring training data. We consider the situation where the learner, with a fixed budget, may 'purchase' data during training. In particular, we examine the case where observing the value of a feature of a training example has an associated cost, and the total cost of all feature values acquired during training must remain less than this fixed budget. This paper compares methods for sequentially choosing which feature value to purchase next, given the budget and user's current knowledge of Naïve Bayes model parameters. Whereas active learning has traditionally focused on myopic (greedy) approaches and uniform/round-robin policies for query selection, this paper shows that such methods are often suboptimal and presents a tractable method for incorporating knowledge of the budget in the information acquisition process.

## 1 Introduction

A recent project was allocated $2 million to develop a diagnostic classifier for cancer subtypes. In the study, a pool of patients with known cancer subtypes was available, as were various diagnostic tests that could be performed, each with an associated cost. Experts theorized that some combination of these tests would be capable of discriminating between subtypes; our challenge was to build a classifier using these tests that would be the most effective.

The first step is to acquire the relevant information: here, we have to decide which tests to perform on which patients. The standard approach, of course, is simple round-robin: run every test on every patient... until we exhaust our fixed budget. Given our finite budget, however, this might not produce the best classifier — e.g., if we can determine that two tests are equivalent, it is clearly inefficient to perform both tests. Fortunately, there are many other options. For example, we could run a subset of the tests on each member of a larger pool of patients, or even decide in a patient-by-patient manner which specific tests to run. Indeed, we could go to the extreme of building a dynamic policy that at each time step decides which tests to perform on which patient, based on all of the information available about the costs and apparent effectiveness of the tests, as well as the remaining available funds.

This paper explores this idea: how to dynamically decide which tests to run on which individual to produce the most effective classifier, subject to the known firm budget.

The rest of this section provides the basic model of our "budgeted learning task," then contrasts this task with many related but distinct ideas. Here we explain in particular how our objective differs from standard bandit problems, on-line learning, active learning, and active classification. Section 2 then provides the foundations: overviewing Naïve Bayes classifiers, (which our learners will return) then the notion of a "policy." Section 3 presents a number of policies, including standard ideas (round-robin and similar) as well as others that are less standard but, as we will see, often more effective. We implemented these systems and ran a number of tests on both real and synthesized datasets; Section 4 reports our findings. We see in particular that round-robin is typically not the most effective policy. The URL [Gre] provides additional information, both theoretical (e.g., proofs) and empirical (datasets, etc.).

### 1.1 Formal Model

As usual, each instance is characterized by a set of $n$ features $\mathbf{X} = \langle X_1, X_2, ..., X_n \rangle$, as well as a class label $Y$. Initially, our learner $R$ knows only the class *labels* of a large set of training instances; n.b., $R$ does not know the *value* of any feature for any instance.

$R$ begins with a known, fixed total budget $b \in \mathbb{R}$, and knows the costs $c_i = c(X_i)$ of each feature. At each time, $R$ can, at cost $c(X_i)$, obtain the value of the $i$-th feature of an instance with a particular label $y$. (Hence, $R$ can explicitly request, say, the $X_1$ value of a $Y = y_1$ instance.) $R$ continues until exhausting its budget. At that point, $R$ returns a classifier. Its goal is to obtain a classifier whose performance is optimal. (Section 2.2 provides more precise definitions of these ideas.)

### 1.2 Related Work

Many on-line learners try to minimize the number of training examples, either explicitly or implicitly (*e.g.*, [MCR93], [SG95], or other PAC results that deal with reducing sample complexity). These approaches, however, allow the learner to acquire as many examples as are needed to meet some requirements — *e.g.*, for some statistical test, or some specified $\epsilon$ and $\delta$ values in the case of PAC-learners [Val84]. We, however, have a firm total budget, specified before the learning begins. Moreover, our approach is fine-grained, as our system can explicitly ask for the value of a *single* specified feature, rather than an entire tuple of values, one for each feature of an instance. (In fact, our results show that this alternative "round-robin" approach is often inferior to other policies.)

This kind of problem is also related to active learning scenarios as described in [TK00], [RM01], and [LMRar]. In typical pool-based active learning, a pool $\mathcal{P}$ consisting of *unlabeled* data instances with completely specified features is available. We are considering the complement of the problem: class labels are available but not feature values. The work in [TK00] could be applied to our problem since it is designed for general belief nets of which our Naïve Bayes model is a special case. However, our goal is to build a good *classifier* as opposed to a good generative model.

In previous active learning results (including [TK00]) greedy methods have been shown effective in reducing training sample size, and deeper lookahead has not been used because of inefficiency and insignificant gains (specifically see [LMRar]). However, we observe that in our case the greedy method often has poor performance, and that looking deeper can pay significant dividends.

Budgeted learning is also related to cost-sensitive learning and active classification (*e.g.*, [Ang92, Tur00, GGR02]), although feature costs in [Tur00, GGR02] refer to costs at *classification* time, while we are con-cerned with cost during the learning phase. Nonetheless, like several active learning results [LMRar, TK00, RM01], we show that *selective querying* can be much more efficient than simplistic methods such as round-robin.

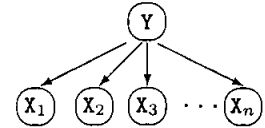

Figure 1: Naïve Bayes Structure

Finally, this work nicely complements the ideas presented in [MLG03], where we consider the budgeted multi-armed bandit problem. That companion paper describes the computational complexity issues surrounding budgeted learning, provides intuition about the behaviour and characteristics of different policies, and details which policies are effective for tackling budgeted learning in the context of multi-armed bandits. While that other work provides a theoretical grounding for budgeted problems in a crisp setting, this paper brings the work of [MLG03] into the realm of full classifier learning.

## 2 PROBLEM STRUCTURE

### 2.1 Naïve Bayes

Our goal is to learn a Naïve Bayes (NB) classifier, which is a belief net with a structure that assumes that feature values are conditionally independent given the class label $y$ [DHS01]; see Figure 1. When presented with a new feature vector $\mathbf{x}$, Bayes' rule is used to compute $\operatorname{argmax}_y P(Y = y | \mathbf{X} = \mathbf{x})$, which is returned as the most likely classification given $\mathbf{x}$ and the model.

### 2.2 Budgeted Learning Formalism

Our learning process must perform a sequence of actions $\{a_{ij}\}$, where action $a_{ij}$ represents requesting the value of feature $X_i$ from an instance with label $Y = y_j$. (*E.g.*, $a_{71}$ represents requesting the value of feature $X_7$ from a $Y = y_1$ instance. Thanks to our Naïve Bayes independence assumptions, it does not matter *which* $Y = y_1$ instance this value comes from.) We can easily view this as a Markov Decision Process [Put94]. We borrow from this framework the notation of *state* $s \in S$ which in our case describes the posterior distribution of each feature $X_i$ given each class label $Y = y_j$ (via $\theta_{ij}$ parameters defined below), *actions* $a_{ij} \in A$ which describe the process of purchasing a feature value to get to a new state, a *reward* function to indicate performance, and a *policy*, which specifies the action to take in each state.

If the feature $X_i$ has $r$ values, then its multinomial parameters $\theta_{ij} = \langle \theta_{ij1}, \ldots, \theta_{ijr} \rangle \sim \text{Dir}(\alpha_{ij1}, \ldots, \alpha_{ijr})$ are Dirichlet distributed, with hyperparameters $\alpha_{ijk} > 0$. E.g., the binary feature $X_7$ associated with the class having its first value $Y = y_1$, might be distributed as $\theta_{7,1,\cdot}^{(s)} = \langle \theta_{7,1,1}^{(s)}, \theta_{7,1,2}^{(s)} \rangle \sim \text{Dir}(3, 8)$.[1] In general, each state $s$ corresponds to the union of these $\alpha_{ijk}$ hyperparameters. If we later request the $X_7$ value of a $Y = y_1$ instance (i.e., take action $a_{7,1}$), and observe the second value $x_{7,2}$, then the posterior distribution in the new state $s'$ will be $\theta_{7,1,\cdot}^{(s')} \sim \text{Dir}(3, 8+1) = \text{Dir}(3, 9)$. In general, if $\theta_{ij}^{(s)} \sim \text{Dir}(\alpha_{ij1}, \ldots, \alpha_{ijr})$, then after observing a value here, which is say the $k$-th value of $X_i$ ($x_{ik}$), the new distribution is $\theta_{ij}^{(s')} \sim \text{Dir}(\alpha_{ij1}, \ldots, \alpha_{ijk}+1, \ldots, \alpha_{ijr})$. Note that the expected value of this $\theta_{7,1,2}^{(s')}$ variable is $\hat{\theta}_{7,1,2}^{(s')} = 9/(3+9) = 3/4$. Given our current knowledge, the probability of incrementing the $k$-th value $\alpha_{ijk}^{(s)}$, when purchasing a value for $X_i$ for the $j$-th label, is $\hat{\theta}_{ijk}^{(s)}$.

The final component of our budgeted learning MDP is the reward function which is only received when the budget has been expended. At that time a reward of $-L(\text{NB}(s))$ is received where $L(\cdot)$ is a loss function of the Naïve Bayes model induced by the parameters of the state $s$. Possible loss functions are described in Section 4, and include such measures as 0/1 error, GINI index, and entropy [HTF01].

## 3 POLICIES

This section first indicates the complexity of finding the *optimal* policy, then outlines a number of plausible policies that we have implemented.

### 3.1 Optimal Policy

Our problem, like all MDPs, has an optimal deterministic policy $\pi^*$ that will result in the maximum expected reward. This policy can in principle be found by any standard MDP solution method (i.e., value iteration, policy iteration, linear programming) in time polynomial in the size of the state space $|S|$. Unfortunately (as is usually the case for any interesting problem) the state space grows exponentially in the number of features, precluding exact computation for problems of an interesting size. In fact, this problem is NP-hard, a result inherited from [MLG03] (see also [Gre]). In light of this, we now examine several simpler, tractable policies that, while not optimal, improve on naïve approaches.

### 3.2 Uniform Policies

Perhaps the simplest policy that immediately comes to mind is a round-robin scheme where features are queried sequentially, regardless of outcomes; see Figure 2(a). If action costs are uniform, this is equivalent to following a *uniform allocation* policy.

Another simple policy when action costs are nonuniform would be to spend $b/|\mathbf{X}|$ on each feature, purchasing more expensive features fewer times. (Hence, ask for around $b/(|\mathbf{X}|\,c(X_i))$ values of each $X_i$.) We call this a *uniform expenditure* policy. We will use these policies as a baseline for empirical performance evaluation.

### 3.3 Biased Robin

Uniform policies are hindered by the fact that they do not respond to the belief states that change as purchases are made. One simple modification to the uniform allocation policy which takes belief change into account is to repeatedly take an action as long as it continues to reduce our current loss. As soon as an action results in an increase in our estimate of the loss, we begin taking the next action. This algorithm is derived from work in [MLG03], and is described in Figure 2(b).

### 3.4 Greedy Loss Reduction

One common technique used in active learning is to calculate the expected loss of taking an action $a_{ij}$ from the current belief state, given by

$$E(\,L(\text{NB}(s))\,|\,a_{ij}\,) = \sum_{s' \in S} P(\,s'\,|\,s, a_{ij}\,)\,L(\text{NB}(s'))$$
(1)

A simple greedy policy is to evaluate these expectations for each possible action from the current state, and perform the action that has the lowest expected loss. This technique has the advantage of directly minimizing the error of the resulting classifier, but we show that this greedy, single-step policy is outperformed by other policies that use deeper lookahead.

### 3.5 Single Feature Lookahead

Although the above greedy method is provably suboptimal, it is tractable since there are typically not too many actions to evaluate. Our goal is to incorporate knowledge of the budget when scoring potential actions while retaining this tractability. We accomplish

---

[1] Here, as $X_7$ is binary, this Dirichlet is also considered a Beta distribution.

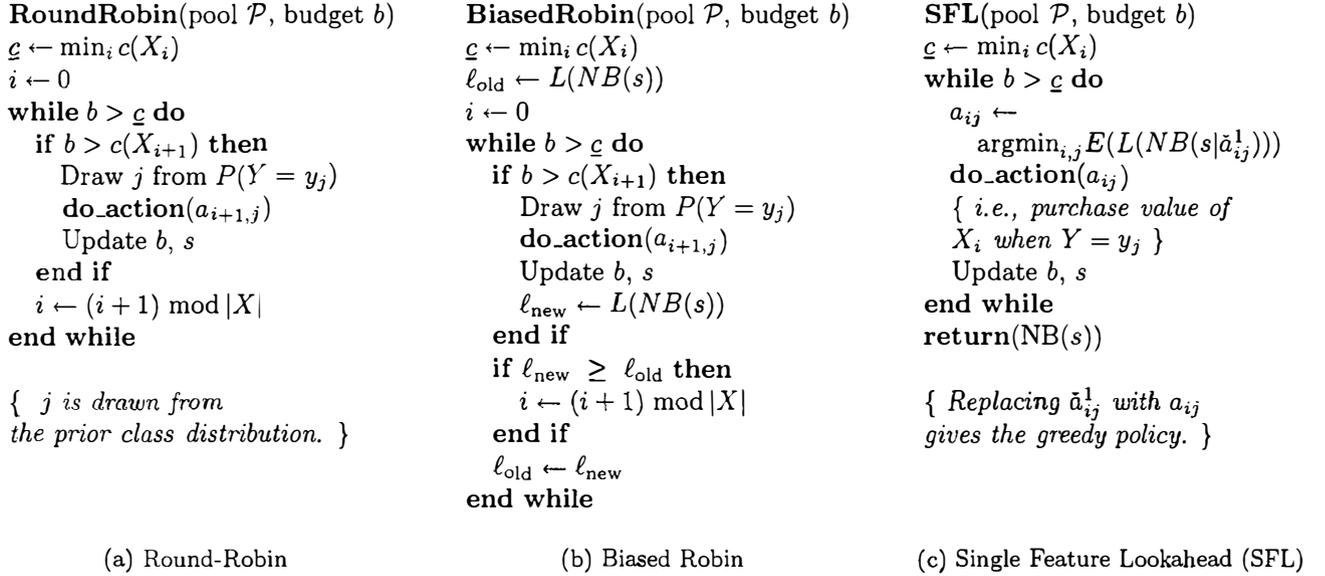

(a) Round-Robin　　　　　　　　　(b) Biased Robin　　　　　　　　(c) Single Feature Lookahead (SFL)

Figure 2: Policies

this by formulating a method that introduces lookahead without expanding the entire state space. (This extends a method from [MLG03].)

An *allocation* is an array of integers that describes the number of times a feature's value is purchased in conjunction with a certain class label; *i.e.*, the number of times action $a_{ij}$ is executed. For example, under allocation $\breve{a}$, $\breve{a}_{32} = 4$ would mean that the value of feature $X_3$ is purchased 4 times from tuples where $Y = y_2$. With normalized uniform costs ($c(X_i) = 1$ for all $X_i$) an allocation can be viewed as an integer *composition*.

The algorithm is based on the ability to calculate the expected loss resulting from performing all actions specified in a static allocation. The expected loss of executing the actions of an allocation $\breve{a}$ is given by

$$E(\,L(\mathrm{NB}(s))\,|\,\breve{a}\,) \;=\; \sum_{s'\in S} P(s'\,|\,s,\breve{a})\,L(\mathrm{NB}(s')) \quad (2)$$

where

$$|\{s' \in S : P(s'\,|\,s,\breve{a}) > 0\}| = \prod_{i,j} \binom{\breve{a}_{ij} + |X_i| - 1}{\breve{a}_{ij}} \quad (3)$$

Again, enumerating all possible states resulting from an allocation is an integer composition problem with a straightforward solution. To simplify things even further, we consider allocations where $\breve{a}_{ij} = \lfloor b/c(a_{ij}) \rfloor$ for some $(i,j)$ and $\breve{a}_{ij} = 0$ elsewhere. We denote this allocation $\breve{a}_{ij}^1$. It represents spending the entire budget on the single action $a_{ij}$, and has a state space of size

$$|\{s' \in S : P(s'|s,\breve{a}_{ij}^1) > 0\}| = \binom{\breve{a}_{ij}^1 + |X_i| - 1}{\breve{a}_{ij}^1} \quad (4)$$

For a feature with two possible values, for example, there are only $2\lfloor b/c(a_{ij}) \rfloor$ distinct states. The probability of reaching any one of these states is given by [Hec95]

$$P(s'|s,\breve{a}_{ij}^1) = \frac{\Gamma(\sum_k \alpha_{ijk}(s))}{\Gamma(\sum_k \alpha_{ijk}(s'))} \prod_k \frac{\Gamma(\alpha_{ijk}(s'))}{\Gamma(\alpha_{ijk}(s))} \quad (5)$$

Our *single feature lookahead* (SFL) operates as follows: For all $i$ and $j$, compute the expected value of the loss of $\breve{a}_{ij}^1$ as defined in (2). We find the action with the minimum loss and perform it **once**, update the belief state and budget, and repeat. The algorithm is described in Figure 2(c).

This policy, like the greedy policy in Section 3.4, is suboptimal. However, the SFL score of an action will be influenced by additional factors that affect the mobility of a distribution. This mobility is affected by a distribution's current belief state (distributions with smaller hyperparameters are more mobile), the cost of an action (the distributions of cheaper actions are more mobile) and by the remaining budget (more budget means more mobility for all distributions.)

## 4　EMPIRICAL RESULTS

We will use the GINI index of the NB classifier as our loss function $L$ for choosing the actions of the greedy

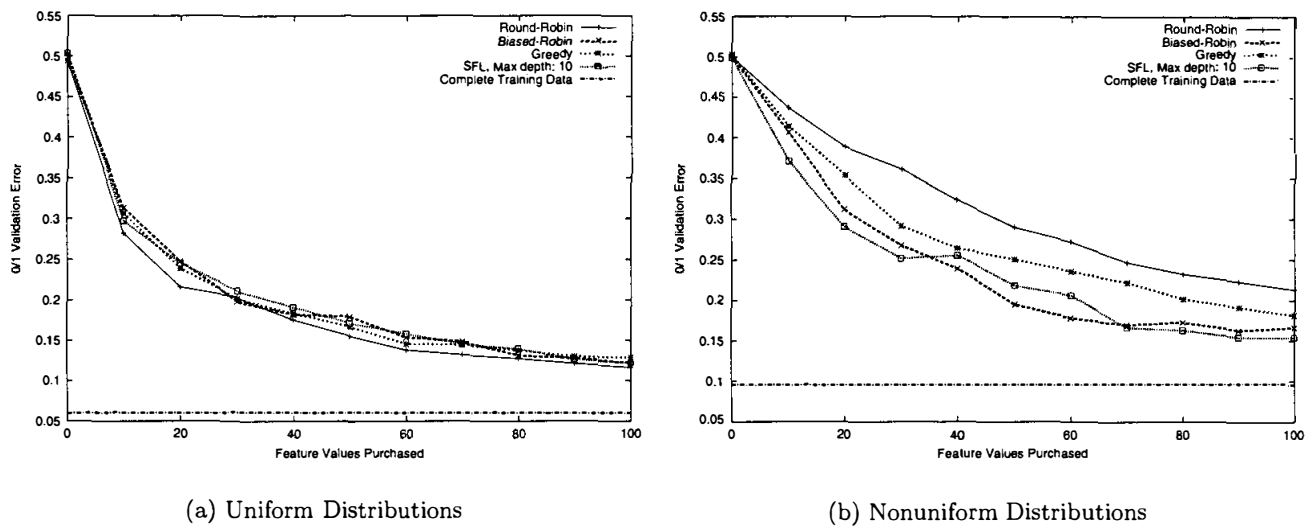

(a) Uniform Distributions  (b) Nonuniform Distributions

Figure 3: Performance on synthesized data. 0/1 error error on a validation set consisting of 20% of the data. Errors are averages of 50 trials.

and SFL policies [HTF01]:

$$L_{\text{GINI}}(\text{NB}(s)) = \sum_{y \in Y} \sum_{\mathbf{x} \in \mathbf{X}} P(\mathbf{x}) P(y|\mathbf{x})(1 - P(y|\mathbf{x}))$$
(6)

We have chosen the GINI index to guide action selection because we found that, being continuous and smooth, it is more sensitive than 0/1 error to the small changes in the NB distribution caused by a single action. (Because 0/1 error is piecewise constant, certain actions would not change its expected value at all.) We have found that both GINI and entropy are similar in this respect.

Although we use the GINI index to choose features, we have plotted 0/1 cross validation error in Figure 3 and Figure 4 (discussed below). This gives a more useful measure of how well the classifiers learned under a budget are performing. In the interest of saving computational time, we have estimated some quantities needed in the greedy and SFL calculations by importance sampling. Because of our NB structure, generating iid samples is trivial. Also, in the case of SFL, we look as deep as either the remaining budget or a fixed maximum indicated by "Max-depth", whichever is smaller. Adjusting this depth allows us to examine the effects of varying degrees of lookahead.

### 4.1 Synthesized Data

Our initial experiments involve data synthesized from Naïve Bayes distributions to test our policies in a setting where the conditional independence assumptions are true. In these experiments, all the feature and class variables are Boolean, with class probabilities at 0.5 each. Each experiment represents an average over 50 trials. In each trial, a Naïve Bayes model with 10 features is generated from defined priors, and the model is used to generate 1000 iid instances, of which the first 80% are used for training and the remainder are used to compute the 0/1 validation error. The vertical axis shows the 0/1 error of the model trained by the various algorithms after a number of purchases. The "Complete Training Data" line is the 0/1 error of the Naïve Bayes model trained on all of the training data.

The two experiments differ in the priors from which the Naïve Bayes model is generated. In the first experiment (Figure 3(a)), under each class, each feature's multinomial parameters are drawn from a uniform distribution; $\langle \theta_{ij\cdot} \rangle \sim \text{Dir}(1,1)$. Therefore, all features are discriminative to varying degrees. We observe that the performances of the algorithms are all comparable, and there is nothing to be gained from selective querying. (i.e., round-robin works well.) The reason for the comparable performance of the algorithms is basically that purchasing any feature is expected to reduce the loss of the whole NB model somewhat, but highly discriminative features are so rare that it does not pay to hunt for them.

Figure 3(b) displays the other extreme where all features except one are irrelevant. In this experiment, each feature's parameters are drawn from a uniform ($\text{Dir}(1,1)$) distribution *independently* of the class. One feature, $X_i$, chosen at random is selected to be discriminative; in particular we set $P(X_i = x_{i1} | Y = y_1) = 0.9$ and $P(X_i = x_{i1} | Y = y_2) = 0.1$. We ob-

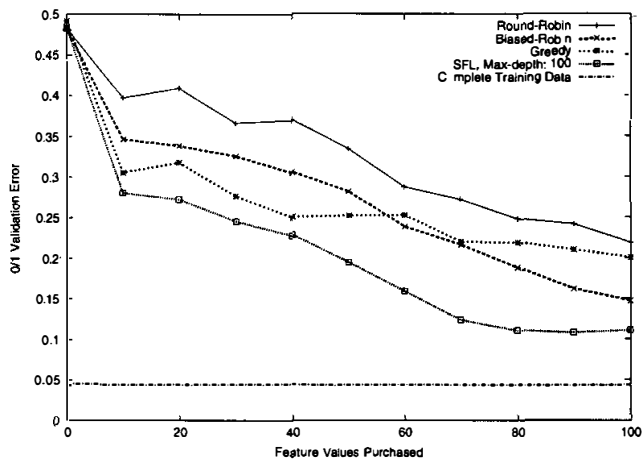

(a) Mushroom Data: Max Depth equal to Budget

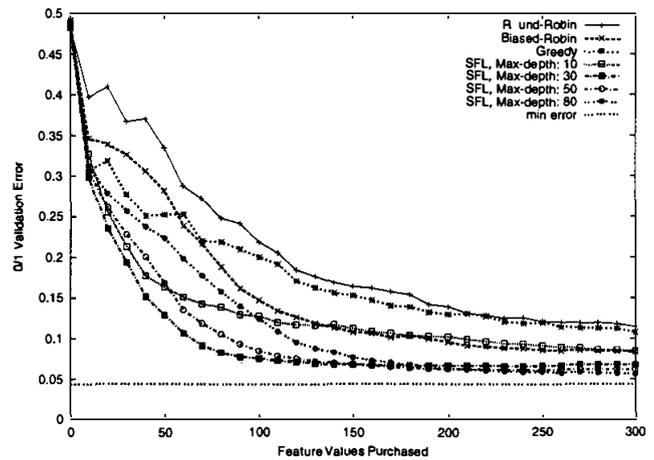

(b) Mushroom Data: Various Max Depths

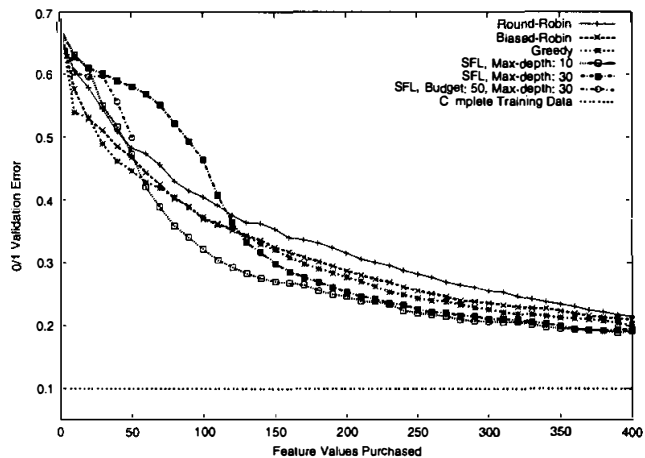

(c) Nursery Data

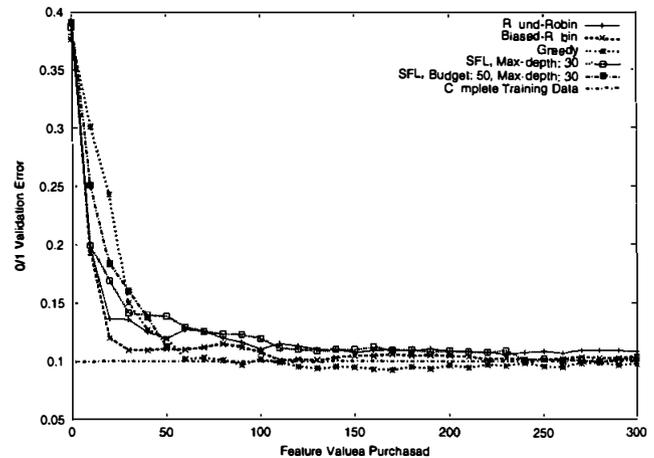

(d) Votes Data

Figure 4: Performance on UCI data. 0/1 error error on a validation set consisting of 20% of the data. Errors are averages of 50 trials.

serve that lookahead (with Max-depth 10) and biased-robin algorithms significantly outperform round-robin in this scenario, requiring only about half as many purchases to obtain the same error level. These policies are capable of identifying the most promising features and obtaining better estimates of their posteriors, which improves performance. We can increase this difference in performance by increasing the discrimination level of the relevant feature (e.g., the extreme case would be $P(X_i = x_{i1} \mid Y = y_1) = 1.0$ and $P(X_i = x_{i1} \mid Y = y_2) = 0$) and by increasing the number of irrelevant features.

## 4.2 UCI DATA

For a less contrived test bed, we have chosen several datasets from the UCI Machine Learning Repository [BM98]. These plots show cross validation error (20% of the dataset) on the mushroom and votes datasets of the different policies. Each point is an average of 50 trials where in each trial a random balanced partition of classes was made for training and validation. The average performance of the Naïve Bayes model trained on the whole training set is also shown ("Complete Training Data").

The mushroom dataset is a binary class problem (poisonous vs. edible), with 22 features, 8124 instances, and a positive class probability of 0.52. One of the features, feature 5, is a very discriminative 10-valued feature, while others are less discriminative [Hol93]. Figures 4(a) and 4(b) show the performance of the different policies.

Figure 4(a) represents what we imagine to be a typical application of the policies discussed in this paper. The budget has been set at 100, and we allow SFL a "Max-depth" of 100, meaning that it always looks ahead as far as its remaining budget. Here we see that the contingent policies (*i.e.*, Biased-Robin, Greedy, and SFL) outperform the simplistic Round-Robin. Of the contingent policies, SFL, which is the only policy to use knowledge of the budget in decision making, performs best.

In Figure 4(b) again we see that the adaptive policies perform best, and we also see the effect of varying the degree of lookahead, with Max-depth 30 SFL dominating earlier in the run and Max-depth 80 SFL performing best later. This plot is illustrative of the effect of altering lookahead depth as it shows 'more greedy' shallow lookahead performing well initially before they are bested by more farsighted policies, indicating that it is important to match the depth of lookahead to the actual budget. One more illustration of this is the difference in performance after 50 purchases between the Max-depth 30 SFL with a budget of 300, (■) and the Max-depth 30 SFL with a budget of 50 (o). As these policies approach the 50 purchase mark, o is not looking beyond a total of 50 purchases, but ■ is still looking 30 purchases ahead at each step, which results in a performance hit. Regardless of depth, SFL is capable of picking out relevant features: Out of the 300 purchases, feature 5 is bought by SFL an average of 75 times, while a non-discriminative feature such as feature 18 is bought an average of only 2 times. For some budgets, the 0/1 error of SFL is nearly half that generated by round robin.

The nursery dataset (Figure 4(c)) is a five class problem with nine features that can take on between two and five values. The relative performances of the policies are closer to each other, but their behaviour is similar to Figure 4(b).

The votes dataset (Figure 4(d)) is a binary class problem (democrat vs. republican), with 16 binary features, 435 instances, and a positive class probability of 0.61. In the votes dataset, there is a high proportion of discriminative features, and we observe that all policies within relatively few purchases reduce the error to the minimum possible for a Naïve Bayes classifier. In fact, because of independence assumption violations, it is possible for selective policies to perform better than a NB classifier trained on the whole data set by doing a form of feature selection. Note that for the SFL policy, the specified budget is significant: if the budget is set at 50, the performance of SFL at 50 purchases is better than its performance at 50 when the budget is set at 300. Other policies do not take the budget into account.

We have observed the same overall patterns on several other datasets that we have tested the policies on so far (CAR, DIABETES, CHESS, BREAST): the performance of SFL is superior or comparable to the performance of other policies, and Biased-Robin is the best algorithm among the budget insensitive policies; see [Gre] for additional details. Run times for round-robin and biased-robin are very short, with the greedy policy taking slightly longer. Run times for SFL took the longest, and were on the order of minutes for all experiments.

## 5 CONCLUSION

### 5.1 Future Work

This work can be extended in several obvious directions. We have chosen to use a Naïve Bayes classifier, but any classifier that can deal with incomplete data tuples could be used, in principle.

An immediate extension of our work is to handle the detection of dependencies among features and dropping redundant features in order to improve the performance of the learned classifier. This can be viewed as a special case of actively learning structure [TK01]. Dependency detection would remedy a problem on some of the UCI datasets such as votes, when a Naïve Bayes classifier using all features performs worse than using only the single best feature for classification. We are currently investigating logistic regression as a means of mitigating the problems caused by unmodeled dependencies.

The cost structure presented here is quite simple, but real data acquisition can have a very complex cost structure. One could imagine for example, extrapolating from the medical study that was our motivation, a situation with a fixed cost for obtaining a new (empty) data tuple with its corresponding class label, followed by incremental feature value costs. (Imagine it costs $50 to have a patient come into a clinic, after which each individual test costs $10.) Concerning the budget, one could consider the scenario where we have a "soft" budget, perhaps with an increasing cost per feature after we have expended our initial $b$. If there are major differences in costs, then the goal should be to learn a *cost-sensitive* or *active* classifier [GGR96, Tur00] still in this budgeted framework.

Though it is too computationally expensive to solve optimally, our problem does have some structure that may be exploitable in its MDP form. We are also interested in the suitability various approximate methods for solving MDPs (*i.e.*, [Duf02]) for use on our problem.

## 5.2 Contributions

In this paper, we have formulated the general "budgeted learning problem" as a Markov Decision Process and shown that its optimal solution appears to require computation time exponential in the number of features. (In fact it is NP-hard.) We have shown that simple policies such as round-robin and greedy loss reduction can be problematic on certain datasets, and propose two alternatives (biased-robin and single feature lookahead) that can perform significantly better than these simple policies in certain situations. Empirical performance results both on synthesized data and on parts of the UCI dataset support our claim that the budget-aware policies we have proposed are preferable when faced with a budgeted learning problem.

### Acknowledgments

Omid Madani was partially supported by an Alberta Ingenuity Associateship. All three authors have been supported in part by NSERC funding and by the Alberta Ingenuity Centre for Machine Learning.